\documentclass[letterpaper, 10 pt, conference]{ieeeconf}  

\IEEEoverridecommandlockouts                              

\usepackage{multirow}
\usepackage{graphics} 
\usepackage{epsfig} 
\usepackage{mathptmx} 
\usepackage{times} 
\usepackage{amsmath} 
\usepackage{amssymb}  
\usepackage[utf8]{inputenc}
\usepackage{subcaption}
\usepackage{pdfpages}
\usepackage{paralist}
\usepackage[font=small,labelfont=bf]{caption}
\usepackage{array}
\usepackage{booktabs}
\usepackage{lipsum}
\usepackage{caption}
\usepackage{siunitx}
\usepackage{filecontents}
\usepackage{hyperref}
\usepackage{float}
\usepackage{cite}

\usepackage{tabularx}
\usepackage{xcolor}

\title{\LARGE \bf
 Can an Embodied Agent Find Your ``Cat-shaped Mug"? \\ LLM-Guided Exploration for Zero-Shot Object Navigation
 \vspace{-8pt}
}

\author{Vishnu Sashank Dorbala$^1$, James F. Mullen Jr$^1$, and Dinesh Manocha$^1$ \\
{\small{Supplemental material including Full Technical Report, Code, Video are at \url{https://gamma.umd.edu/LGX/}}}
\vspace{-10pt}
\thanks{
$^1$The authors are associated with the University of Maryland, College Park, USA
{\tt\small vdorbala@umd.edu, mullenj@umd.edu, dmanocha@umd.edu}}%
}
\newcolumntype{M}[1]{>{\centering\arraybackslash}m{#1}}

\begin{document}
\maketitle
\thispagestyle{empty}
\pagestyle{empty}

\begin{abstract} 
We present LGX (Language-guided Exploration), a novel algorithm for \textit{Language-Driven Zero-Shot Object Goal Navigation} (L-ZSON), where an embodied agent navigates to an \textit{uniquely described} target object in a \textit{previously unseen} environment. Our approach makes use of Large Language Models (LLMs) for this task by leveraging the LLM's commonsense-reasoning capabilities for making sequential navigational decisions. Simultaneously, we perform generalized target object detection using a pre-trained Vision-Language  grounding model. We achieve state-of-the-art zero-shot object navigation results on RoboTHOR with a success rate (SR) improvement of over 27\% over the current baseline of the OWL-ViT CLIP on Wheels (OWL CoW). Furthermore, we study the usage of LLMs for robot navigation and present an analysis of various prompting strategies affecting the model output. Finally, we showcase the benefits of our approach via \textit{real-world} experiments that indicate the superior performance of LGX in detecting and navigating to visually unique objects.
\end{abstract}

\section{Introduction}
 Humans do not conform to preset class labels when referring to objects, instead describing them with free-flowing natural language. Robot agents performing \textit{object goal navigation} in household environments must be able to comprehend and efficiently navigate to this seemingly infinite, arbitrary set of objects defined using natural language. For instance, a human may ask the robot agent to find its ``cat-shaped mug." An agent trained on rigid class labels may interpret this as the human asking for a ``cat" or a ``mug" when the human is really referring to a mug in the shape of a cat. These types of unique objects typically lie outside the domain of the object categories commonly found in large image datasets such as ImageNet 21k \cite{dengImageNetLargescaleHierarchical2009} and OpenImages V4 \cite{openimages}. Additionally, agents deployed in household environments may be required to navigate to these target objects without explicitly having a map or layout of the house available.

In our work, we aim to address these issues by tackling the \textit{L-ZSON} task \cite{gadreCoWsPastureBaselines2022}. \textit{L-ZSON} or Language-Driven Zero-Shot Object Navigation involves the agent using a \textit{freeform natural language description} of an object and finding it in a ``zero-shot" manner, without ever having seen the environment \textit{nor} the target object beforehand. 
The conventional Object Goal Navigation task \cite{objg3, objg4} requires the agent to locate an object from a predetermined category set within an unseen environment. Zero-Shot Object Navigation (ZSON) \cite{objg1, objg2} extends this task to identify a target without any environment-specific training. \textit{L-ZSON} further extends ZSON to freeform natural language descriptions.

 \begin{figure}[t]
     \centering
     \includegraphics[width= \linewidth]{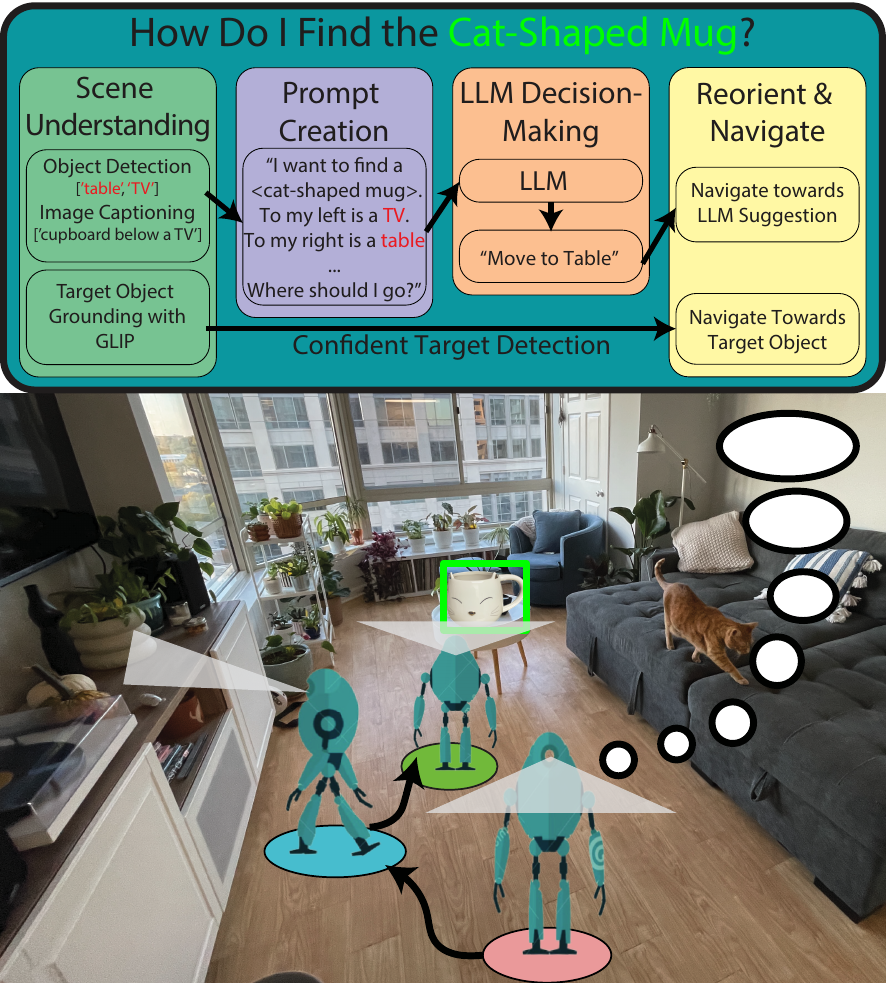}
     \caption{\textbf{LLM-Based Navigation}: Our method, LGX approaches the problem of Language-driven Zero-Shot Object Navigation or L-ZSON. To navigate to and detect an unseen, arbitrarily described object class in an unknown environment, we first extract visual semantic information about the environment. This information is utilized to develop a prompt for the Large Language Model (LLM), whose output provides us with either object sub-goals or cartesian directions to guide the embodied agent towards the target. Meanwhile, GLIP searches for the environment for the target object, which in this case is a ``\textit{cat-shaped mug}".
     }
     \label{fig:coverimg}
     \vspace{-0.6cm}
 \end{figure}
 
Simulation environments for object navigation tasks, including RoboTHOR \cite{deitkeRoboTHOROpenSimulationtoReal2020} and AI Habitat \cite{savvaHabitatPlatformEmbodied2019}, only contain common day-to-day household objects described using simple language (eg. Mug, Table, Bed). However, humans tend to use unconstrained, natural language when talking to agents \cite{uncons1}, leading to potential confusion in the agents' interpretation \cite{ambiguity1}. This problem becomes more apparent in the sim2real transfer of common object navigation models \cite{sim2real1} that fail to understand the humans instructions. In this work, we seek to address this issue by carrying out real-world experiments with unique object references (eg. olive-colored jacket).

Common approaches to solving Object Goal Navigation are based on fully supervised learning \cite{objg3, duet}, which is not practical for an agent that is expected to detect arbitrarily described objects and perform consistently in dynamic real-world environments.
While some recent works address generalizability to new locations via the ZSON task \cite{zson}, even fewer address the issue of generalizing to novel objects \cite{CoW} with the L-ZSON task, and none study real-world test cases that contain an abundance of unconstrained language. \\
These works utilize large-scale pre-trained models such as CLIP \cite{CLIP} and GLIP \cite{liGroundedLanguageImagePretraining2022} to perform zero-shot open-vocabulary object detection in the wild. 
The downstream transfer of such `foundation models' \cite{foundation} has shown great improvement in various vision and language tasks such as image captioning \cite{clipcaption} and question answering \cite{gptuse}. This transfer to robotics is non-trivial however, as unlike vision and language, robot tasks usually involve some form of \textit{experiential} decision-making as the agent continuously interacts with the environment. Exploiting the implicit knowledge contained by these models to compose robot actions presents a unique challenge.

\noindent {\bf Main Contributions:}
Motivated by the challenges above, we present, \textbf{LGX} or \textit{Language-Guided Exploration}, a novel approach that leverages the implicit knowledge of large language models (LLMs) and pre-trained vision and language models to tackle the L-ZSON task.

Object goal navigation including L-ZSON can be broken down into two key components --- \textit{Sequential Decision Making} and \textit{Target Object Grounding}. The former refers to making exploratory decisions on the go, while the latter refers to locating and \textit{grounding} a target object from agent percepts.

In this work, we make use of Large Language Models (LLMs) and Vision-Language (VL) Models to address generalizability issues that hinder the performance of both these components. As LLM's rely on the prompts being used \cite{LLMprompt}, we study the influence of prompt formulation via in-context learning \cite{gpt3} and present a case-by-case analysis of the effect of various prompt types. Additionally, we analyze the usage of VL models for Target Object Grounding and show improved performance with unique object references.
We make the following contributions:

\begin{figure*}[t!]
    \centering
    \includegraphics[width=\linewidth]{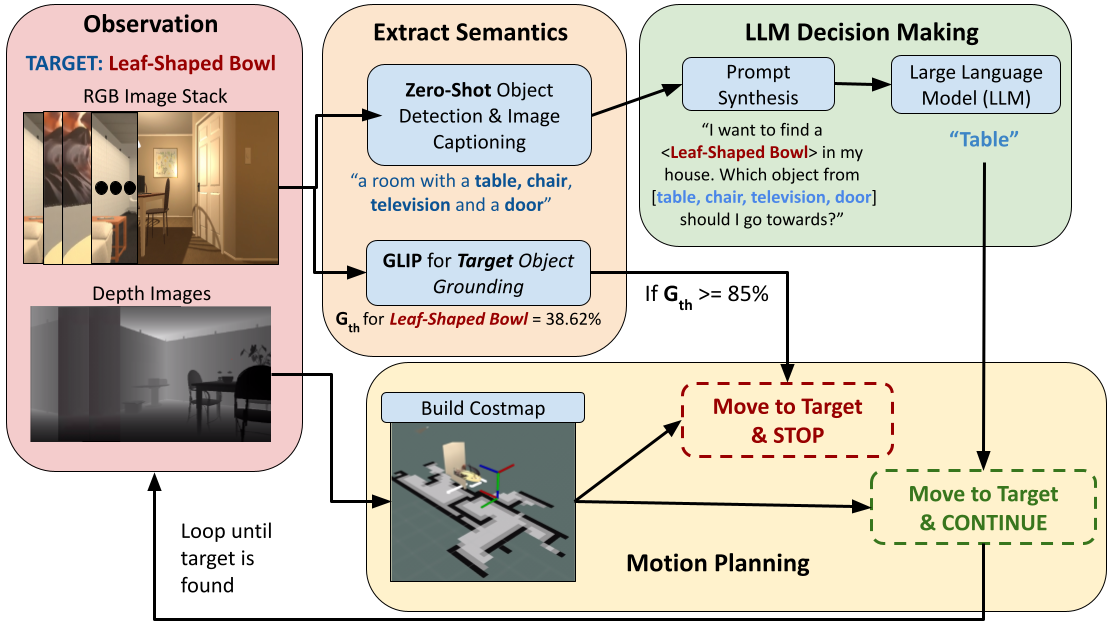}
    \caption{An overview of our approach. We first gather observational data from the environment by performing a 360 degree rotation to obtain depth and RGB images around the agent. The RGB images give us semantic information about the objects in the agent's view, while the depth image allows us to create a costmap. We then synthesize prompts for the LLM by utilizing the extracted object labels. Finally, the LLM drives the navigational scheme by producing an output from the object list, which tells the agent which direction to head towards. Simultaneously, we attempt to ground the target object in the scene with GLIP. When the target is found, we exit the decision making loop and navigate directly to it.
    }
    \label{fig:LLMNav}
    \vspace{-0.5cm}
\end{figure*}

\begin{enumerate}
\item We present LGX, a novel approach to tackle L-ZSON, a language-guided zero-shot object goal navigation task. Our approach localizes objects described by unconstrained language by making use of large-scale Vision-Language (VL) models and leverages semantic connections between objects built into Large Language Models (LLMs). Specifically, we study the implicit commonsense-reasoning capabilities of LLMs in assisting the sequential navigational decisions necessary to perform zero-shot object navigation.
\item Our approach utilizes visual scene descriptions of the environment to \textit{formulate prompts} for LLM's, the output of which drives our navigation scheme. We study various types of prompts and provide insights into successfully using these prompts for robot navigation.
\item Our approach shows a 27\% improvement on the state-of-the-art zero-shot success rate (SR) and success weighted by path length (SPL) on RoboTHOR.
\item Finally, we also present a transfer of our method onto a real-world robotics platform and study the various complexities involved in this setting. To the best of our knowledge, ours is the first approach to evaluate the performance of L-ZSON methods in the real world.
\end{enumerate}

\section{Related Work}
\subsection{Language-Guided Robotics}
Using language to guide robots is a popular task in literature, with work ranging from using generalized grounding graphs \cite{tellex2} for robot manipulation \cite{ggs} to performing language-guided navigation \cite{safenav, robotrust}.
Thomas et. al in \cite{tellex3} presents an approach to parse unconstrained natural language via a systematic probabilistic graph-based approach. More recent work tackling this problem by Jesse et. al. \cite{jesse1, cvdn} and Gao et. al. \cite{dialfred} has explored the use of human-robot dialogue to gather relevant information for completing tasks. Parsing unconstrained natural language is very relevant in our work, and we are motivated by the techniques developed by these papers.

\subsection{Language-Driven Zero-Shot Navigation}
Recent works have attempted to use CLIP \cite{CLIP} for performing zero-shot embodied navigation. CLIP is a large pre-trained Vision-Language model that is capable of zero-shot object detection.  
Dorbala et. al. in \cite{clipnav} use CLIP to perform Vision-and-Language navigation in a zero-shot manner, while Gadre et.al in \cite{CoW} have used it to perform object goal navigation. Both these works work under the assumption of unseen environments.

L-ZSON introduced by Gadre et. al in \cite{gadreCoWsPastureBaselines2022} approaches the problem of zero-shot object navigation, using uncommon target objects. They obtain a baseline for this task using OWL-ViT, a finetuned vision transformer for object grounding, and frontier-based exploration (FBE) \cite{FBE}. In contrast, our approach uses GLIP \cite{liGroundedLanguageImagePretraining2022}, a pre-trained VL model for zero-shot object grounding. To explore the environment, we incorporate GPT-3 \cite{gpt3}, an LLM, to make navigational decisions.

\subsection{Language-Guided Scene Manipulation}

Language-guided scene manipulation is a popular task, requiring agents to process natural language instructions to manipulate objects in the environment. The authors of \cite{bugardlearning1-rev2, bugardcomposing-rev3, asfourscene-rev5} present various semantic-segmentation-based approaches for scene manipulation and object placement using language prompts as input. There also exist a broad range of classical approaches for language-guided manipulation \cite{ggs, langexectutionjacob-rev11, tellexnlp-rev12, groundedjacob-rev13, guarenteesraman-rev14} that utilize different forms of grounding graphs to perform this task. Raman et. al in \cite{sorrydaveraman-rev10} tackle a unique manipulation task, synthesizing language that the robot can use to \textit{explain} its failure to the user.

More recently, several works have presented novel approaches for language-conditioned manipulation task using pre-trained embeddings \cite{zhaoplacement-rev6, shridhar2022cliport-rev7, manipulation-conditioning-rev8, languageconditioning-rev9}. Our work focuses on language-guided exploration or LGX, as opposed to these manipulation tasks.

\subsection{LLMs for High-Level Task Planning}

Recent works have established LLM's as a tool for creating plans for the agent to execute, in breaking down high level instructions (such as \textit{``Make me breakfast"}) to lower-level tasks for the agent to perform (such as \textit{``Make eggs"} and \textit{``Bring juice"}) \cite{abbeelplanners-rev15, saycan-rev16, innermolologue-rev17}. Note that these approaches utilize it for generating and planning actions for the robot, but not explicitly for \textit{\textbf{exploration}} in the wild. Information about the environment is already assumed to be known, and the LLM is used for breaking down complex tasks for the robot to execute.\\

Huang et al. in \cite{abbeelplanners-rev15} looks at using an LLM to simplify complex tasks into ``actionable sub-tasks", for a robot then act upon. Their task involves using pre-trained LLM's at a \textit{planning level} in an interactible environment to determine how to interact with a known VirtualHome environment, and does not use the outcome to explore it.\\

SayCan in \cite{saycan-rev16} utilizes an LLM in a similar manner to deduce actions that the agent can take, but additionally also uses an affordance function to ground the generated actions to the environment. They assume the object locations in the map to be known (see section D.2 in said paper) and solely tackle the ``planning problem", with pick, go to and place actions already available. They do not consider a zero-shot exploration case like LGX to navigate in an unseen environment.\\

Huang, Xia, and Xiao et al., in \cite{innermolologue-rev17}, also follow a similar \textit{planning} objective and utilize the LLM in tandem with human feedback to accomplish the instruction.\\

In contrast, our work uses an LLM as a prior not for planning, but rather to explore an unseen environment to predict the direction that the robot needs to take without requiring a map.

\subsection{LLMs for Language-Guided Navigation}
The adaptation of Large Language Models in robotics has recently been garnering interest.
Several recent works \cite{abbeelplanners-rev15, saycan-rev16, innermolologue-rev17}
have also used LLMs specifically for their planning capabilities. 
Note we are different from these works as they utilize the LLM for \textit{planning objectives} while we utilize the LLM for \textit{environment exploration.}

A few works have looked at using generative models for navigation, specifically, LM-Nav \cite{shahLMNavRoboticNavigation2022} and VLMaps \cite{huangVisualLanguageMaps2022}. Both these works look at solving the Vision-and-Language Navigation (VLN) problem, where the input is the unconstrained language describing a \textit{path to the goal}. The latter uses GPT-3 \cite{gpt3} to obtain ``landmarks" or subgoals, while the former focuses on using an LLM for ``code-writing" \cite{progprompt}.
In contrast, our focus is on translating visual scene semantics into input prompts for the LLM to obtain navigational guidance in the form of actions. We directly incorporate the LLM output into a sequential decision-making pipeline to drive our agent's navigation scheme.


\section{Solving L-ZSON using Language-Guided Exploration (LGX)}

\subsection{Method Overview}


We present an overview of our method in Figure \ref{fig:LLMNav}. Our approach uses a Large Language Model (LLM) to predict where the agent needs to navigate. To do this, we first extract contextual cues from the scene in the form of object labels or scene captions. Either of these cues are then used to devise a prompt asking the LLM about which how the agent should proceed to explore the environment. The LLM uses its commonsense knowledge about object relationships in the environment to provide the agent with a direction or object for it to move towards.

Simultaneously, we use a Vision-Language model, GLIP \cite{liGroundedLanguageImagePretraining2022} to obtain a target object grounding score, which gives us the confidence of the described target being in the scene. Once the confidence meets a threshold $G_{th}$, the agent assumes the target object to be in its egocentric view. An episode is rendered successful if the target object is in the agent's view while performing rotate-in-place.

\subsection{Scene Understanding}
\label{sec:scene_understanding}
In each run, the agent observes the environment, gathering RGB and depth images for inspection. During each observation, we have the robot rotate in place 360 degrees, taking images at a set resolution $r$.
This leaves us with $360/r$ RGB images, $I_r$, and depth images $I_d$. $I_d$ is used to construct a 2D costmap of the environment. Every image in $I_r$ is then fed into either an object detection or an image captioning model. Both these models give us different results, which we discuss in the experimentation section. For object detection, we use YOLO \cite{redmonYouOnlyLook2016}, which contains common household classes, while BLIP \cite{BLIP} gives us image captions. 
BLIP produces descriptive captions $C$ of the environment, while YOLO gives us a list of objects $O$ around the agent that it can potentially navigate towards. We chose either $C$ or $O$ to as part of our prompt to the LLM.

The rotate-in-place at each step allows the agent to fully observe its surroundings, giving the LLM enough information to make a fully informed navigation decision from the agent's current position in the environment. Without it, the agent would proceed toward seen objects over unknown space, even if none of the seen objects were related to the goal object, $o_g$. For example, if $o_g$ is a "blue pillow," but it is initialized facing a kitchen and we see objects such as "microwave," "mug," and "table," the robot will proceed to explore near those objects because it does not know that directly behind it is a "bed" or a "couch", which is potentially where the pillow might be.

Simultaneously, while performing the full circle rotation, the agent uses $I_d$ to construct a costmap of the environment. We use RTABMAP \cite{rtabmap} that uses visual correspondences along with depth information from the standard costmap\_2d ROS package \cite{ros}, to compute the costmap. Once a navigational decision is made by the LLM by providing either an object or a direction (depending on if $C$ or $O$ is passed to the input), we reorient the agent accordingly and randomly choose a point in the cost map along the agent's egocentric field of view. The costmap allows us to avoid obstacles while exploring the environment.

\noindent We use GLIP for target object grounding. During each rotation step the agent takes, the collected RGB images are passed through GLIP along with the target object as a prompt. When the grounding accuracy of GLIP is beyond a threshold $G_{th}$, we assume that the target object is in view of the agent, which triggers a STOP signal. If not, the agent continues exploring till $n_r$ number of rotate-in-place turns. The episode is rendered successful if the ground truth target object lies in the view.

$G_{th}$ and $n_r$ are hyperparameters that are empirically chosen from ablation experiments. For selecting $G_{th}$, we ablate with various threshold values in an environment, picking the one with the highest success rate (refer Table \ref{tab:gthval}). $n_r$ is chosen based on the size of the environment.

\subsection{Intelligent Exploration with Large Language Models}
\label{sec:exploration}
We utilize the extracted semantics to devise a prompt for our LLM, GPT-3. 
There are two scenarios we explore,
\begin{enumerate}
    \item \textbf{YOLO $\rightarrow$ LLM }: In this case, we utilize the list of objects $O$ that YOLO detects around the agent to synthesize a prompt. For improving object detection, we set the rotate resolution $r$ to a lower value here.
    \item \textbf{BLIP $\rightarrow$ LLM}: Here, we utilize image captions $C$ generated from the previous step to create a prompt. The rotate resolution $r$ is set to $90$ here, referring to either of the 4 directions (Left, Right, Front, Back) that the agent can take while exploring.
\end{enumerate}






The LLM output upon using the YOLO + LLM approach gives us an object from $O$ to navigate towards. The agent then reorients itself towards this object.
While using BLIP + LLM, the output gives us a direction towards which the agent reorients itself.
When the LLM output does not follow these expected outcomes, the agent chooses a random direction.


\subsection{Goal Detection and Motion Planning}
\label{sec:goaldetandmp}
While completing the $360^{\circ}$ rotation-in-place, we also run GLIP with the goal object $o_g$ as its target label. 
Should GLIP find the target object $o_g$ with a high enough confidence $G_{th}$, we terminate the exploration loop with the target in front of the agent.
We then use the simulator as a ground truth to identify if the target actually lies in front of the agent, thus determining our success case.
If GLIP does not find the target, we continue the LLM-based exploration.

At each step, we first orient the agent based on the LLM's decision. We then use the constructed cost map (refer section \ref{sec:scene_understanding}) to pick a point at a fixed exploratory distance $e_d$ in the direction that the agent is facing. We then use the standard ROS \textit{move\_base} package to avoid obstacles.



\section{Analyzing our Approach}

\subsection{Using GLIP for Zero-Shot Detection}
Open-vocabulary (OV) grounding models have demonstrated strong zero-shot performance to object-level recognition tasks and proved to generalize well across numerous data sources. 
In our approach we use GLIP for grounding target objects of interest. It gives us a bounding box that allows us to localize a direction for navigation in the agent's field of view. GLIP outputs can be defined as

\begin{equation} \label{eq:glip}
    \{o_{t,i}, b_{t,i}\} = GLIP(I_t, P_o)
\end{equation}

where $o_{t,i}$ and $b_{t,i}$ are the object detections and bounding boxes respectively. $I_t$ and $P_o$ represent the input image and the input prompt, defining the objects of interest, respectively.

\begin{figure}[t]
     \centering
     \includegraphics[width=\linewidth]{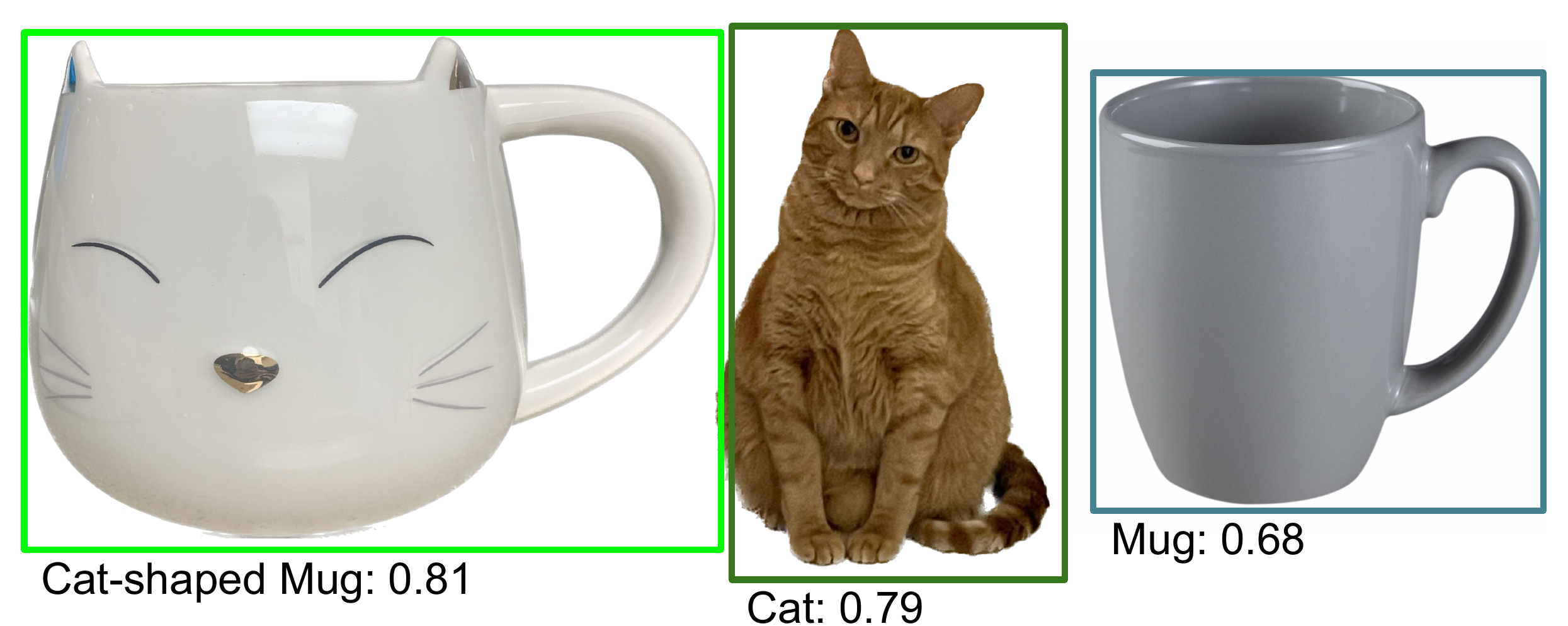}
     \caption{An example of GLIP output when fed with the input string ``Cat-shaped mug . Cat . Mug" on the image given. GLIP can successfully locate a unique object, like a ``cat-shaped mug" and differentiate between it and related objects like a cat or a mug. 
     }
     \label{fig:glip}
     \vspace{-0.5cm}
 \end{figure}

We chose GLIP as our grounding model, gauging its outcome to be superior after ablation experiments with other state-of-the-art OV detection models including OWL-ViT \cite{owlvit} and Object-Centric OVD \cite{ocovd} and Detectron2 \cite{detectron2}.
An example of the usefulness of GLIP for L-ZSON can be found in figure \ref{fig:glip}. It can not only identify the object defined using natural language, the ``Cat-shaped mug", but differentiate it from related objects. Because of this behavior, running GLIP during our rotate-in-place procedure allows us to confidently detect our goal objects $o_g$ irrespective of how they are described.

\subsection{Examining LLM Prompts for Exploration}

The outcome of GPT-3 is greatly influenced by the prompts that it is given. Since we directly use its commonsense reasoning capabilities (a \textit{cat-shaped mug} is more likely to be near a table than a bed) for navigation, it is important for us to consider various prompting strategies.
Should the LLM not provide us with a valid response, the agent moves in a random direction.
We compare seven different LLM prompts which are variations of the following template :

\begin{quote}
    ``\textit{You are controlling a home robot. The robot wants to find a $o_g$ in my house. Which object from $\{O\}$ should the robot go towards? Reply with ONE object from the list of objects.}"
\end{quote}
 We first explore how different points of view LLM feedback. These prompts are below:
\begin{itemize}
  
    \item \textbf{Robot-Prompt}: ``\textit{You are controlling a home robot. The robot wants to find a $o_g$ in my house. Which object from $\{O\}$ should the robot go towards? Reply with ONE object from the list of objects.}"
    \item \textbf{I-Prompt}: ``\textit{I want to find a $o_g$ in my house. Which object from $\{O\}$ should I go towards? Reply in ONE word.}"
    \item \textbf{Third-Person-Prompt}: ``\textit{A $o_g$ is in a house. Which object from $\{O\}$ is likely closest to $o_g$? Reply with ONE object from the list of objects.}"

\end{itemize}

Second, we vary the order of the information given to the prompt.
\begin{itemize}
    \item \textbf{$\{O\}$-First-Prompt}: ``\textit{You are controlling a home robot. You must select one object from $\{O\}$ that the robot should go towards to try to find $o_g$ in my house. Reply with ONE object from the list of objects.}"
    \item \textbf{Get-Closest-Prompt}: ``\textit{You are controlling a home robot. The robot wants to find a $o_g$ in my house. Which object from $\{O\}$ is probably the closest to $o_g$? Reply with ONE object from the list of objects.}"
    \item \textbf{``ONE word"-First-Prompt}: ``\textit{Reply with ONE word. You are controlling a home robot. The robot wants to find a $o_g$ in my house. Which object from $\{O\}$ should the robot go towards?}"
\end{itemize}

Last, we create prompts with natural language captions of the scene. 

\begin{itemize}
    \item \textbf{BLIP-Prompt}: ``\textit{I want to find a $o_g$ in my house. In Front of you there is $<$caption$>$. To your Right, there is $<$caption$>$. Behind you there is $<$caption$>$. To your Left there is $<$caption$>$. Which direction from {Front, Right, Behind, Left} should I go towards? Reply in ONE word.}"
\end{itemize}
\section{Experiments and Results}
\subsection{Experiment Setup}
\textbf{Simulation Setup.} We use the RoboTHOR \cite{deitkeRoboTHOROpenSimulationtoReal2020} validation set as a simulation environment for our experiments. It contains 1800 validation episodes with 15 validation environments. 12 different goal object categories are present. 
Each exploratory turn carries out the scene understanding procedure described earlier in sections \ref{sec:scene_understanding} and \ref{sec:exploration}.
For each episode, we run LGX for $n_r$ exploratory turns or until it detects the target object above the $G_{th}$ threshold. These constraints form the \textit{STOP} condition. $n_r$ is set to 5, given the small environments in RoboTHOR, where the target object is usually within 10 meters of the spawning point. $G_{th}$ is set to $0.85$ for RoboTHOR after ablation experiments described in Table \ref{tab:gthval}. The exploratory distance $e_d$ described in \ref{sec:goaldetandmp} is set to $5$m, ensuring significant changes to the scenery in RoboTHOR after motion.


\begin{table}[h!]
    \centering
    \begin{tabular}{|c|c|c|} \hline 
         $G_{th}$&  SR (\%)& SPL (\%)\\ \hline 
         0.6& 13.8& 7.2\\ \hline 
         0.75&  20.3& 10.8\\ \hline 
         0.8&  32.5& 18.7\\ \hline 
         \textbf{0.85}&  \textbf{35.0}& \textbf{21.9}\\ \hline 
         0.95&  18.0& 11.3\\ \hline
    \end{tabular}
    \caption{Ablations on $G_{th}$ on the RoboThor Validation Set: $G_{th}$ is thresholded using empirical evidence from ablations. A low value produces many false positives, leading to poor performance. Conversely, a high value rejects many potentially successful cases.}
    \vspace{-0.3cm}
    \label{tab:gthval}
\end{table}


\textbf{Prompt Selection Setup.}
We run each of the 7 prompts described in the previous section on a subset of the 500 best and worst performing episodes.

\textbf{Metrics.} We report and compare Success Rate (SR) and Success Rate weighted by inverse path length (SPL) \cite{embodiednav2}. SR and SPL are the primary metrics used in both the Habitat and RoboTHOR challenges. For our prompt ablations, we define a new metric, \textbf{Prompt Success Rate (PSR)} as:
\begin{equation}
    PSR = \frac{p_{suc}}{p_{total}}
\end{equation}
 where $p_{suc}$ denotes the number of instances where the LLM chooses a valid response, and $p_{total}$ denotes the total number of times the agent prompts the LLM. A valid LLM response is when it chooses either an object detected by the agent or a direction for navigation, depending on the semantic extraction scheme used.

\textbf{Real World Setup.}
We conduct experiments with a TurtleBot 2 to validate two facets of LGX in the real world --- i) the \textbf{LLM's Exploration Capability} and ii) the \textbf{GLIP-based open-vocabulary Grounding}.

\begin{figure}[t]
     \centering
     \includegraphics[width=\linewidth]{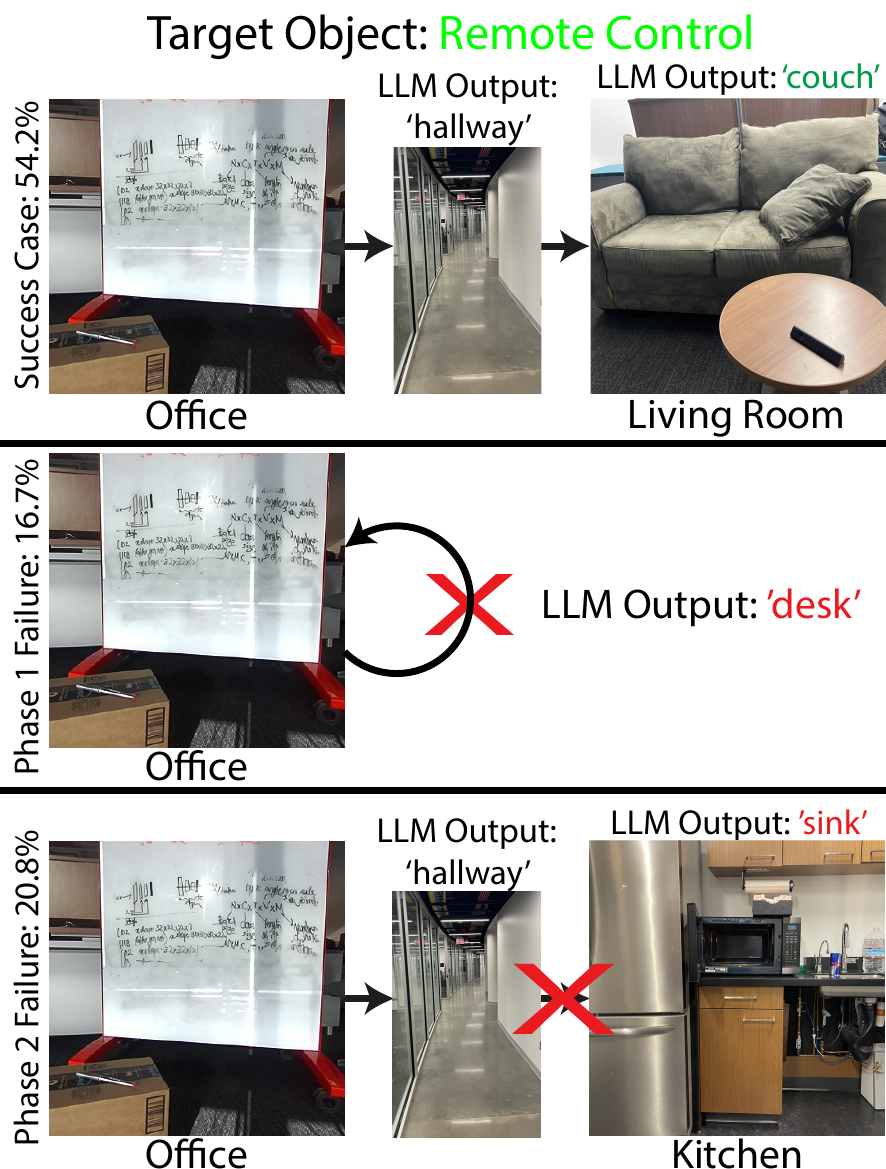}
     \caption{To validate the LLM's exploration capability, we define a two-phase process. The target object is present in a different room, requiring the agent to navigate out of the current room into a \textit{`hallway'}. The LLM in LGX takes objects in the current room along with the hallway as input to the LLM. Not reaching the `hallway' is a \textbf{\textit{Phase 1}} failure. For \textbf{\textit{Phase 2}}, four possible rooms are visible and the agent must navigate to the room with the goal object.
     We pass a set of common objects for each room as shown in Table \ref{tab:real-layouts} as input to the LLM in LGX. Not choosing the correct room is considered a \textit{Phase 2} failure. }
     \label{fig:real-world-gpt}
     \vspace{-0.5cm}
 \end{figure}

To validate i), we look at a \textbf{two-phase approach} where the agent is required to travel from one room through a `hallway' to reach a room containing the target object. In \textbf{Phase 1}, the agent performs rotate-in-place in the spawned room gathering information about objects around it. Since the target object is not present in this room, the LLM-output is expected to be \textit{`hallway'}.
In \textbf{Phase 2}, the agent is present in the hallway and is expected to choose the correct room to navigate to, given a set of common objects found in them (Refer Table \ref{tab:real-layouts}). The LLM-output is now expected to point towards the room that is most likely to contain the target object, based on commonsense knowledge (`remote control' near to `couch'). This is explained in detail in figure \ref{fig:real-world-gpt}.

To validate ii), we examine GLIP's accuracy in classifying unique household object classes. In order to do this, we first define and pick unique target objects, as well as common objects belonging to different household rooms. These are shown in Table \ref{tab:real-layouts}. A success case is defined by a successful GLIP detection of the target object.

\begin{table}[h!] 
    \centering
    \begin{tabularx}{\columnwidth}{l|c|c} 
        Room & Target Objects & Common Objects\\ [0.5ex] 
        \hline\hline
        Kitchen & Red Bull can, Stevia sugar packets & sink, fridge\\
        \hline
        Living Room & remote control, coffee table & couch, tv \\
        \hline
        Bedroom & bust, olive-colored jacket & bed, blanket\\
        \hline
        Office & silver pen, whiteboard & desk, computer\\
        \hline
    \end{tabularx}
    \vspace{5.0px}
    \caption{\textbf{Object Setup for validating LLM Exploration.} We define four household rooms populated with common and target objects that are likely to be found in them. Common objects are regular household items, while target objects are uniquely described with free-form language.}
    \label{tab:real-layouts}
    \vspace{-0.3cm}
\end{table}

This experimental setup validates our system against two main complicating factors of the real world, \textit{free-form natural language}, and \textit{partially-observable environments}. The \textit{free-form natural language} problem is addressed through unique descriptions of each target object, which are not common visual class labels. The \textit{partially-observable environments} component is addressed by conducting a two-phase exploration experiment, where only a few objects are visible in each room, replicating real-world homes.

\subsection{Baselines and Ablations}
We compare our method, LGX with two state-of-the-art methods and an ablative method:

\textbf{CLIP-on-Wheels (CoW).} \cite{gadreCoWsPastureBaselines2022} use Grad-CAM, a gradient-based visualization technique with CLIP\cite{CLIP} to localize a goal object in the egocentric view. CoW employs a Frontier-based Exploration technique for zero-shot object navigation.

\textbf{OWL CoW.} \cite{gadreCoWsPastureBaselines2022} utilizes the OWL-ViT transformer, in place of a CLIP model for target object grounding. This detector then replaces CLIP in the CoW method.

\textbf{GLIP on Wheels (GoW).} Where \cite{gadreCoWsPastureBaselines2022} utilizes the OWL-ViT transformer for its visual object grounding, we replace it with our GLIP based grounding system. This is also an ablation of our method without our LLM-based exploration mechanism.

\textbf{Random with GLIP.} As a baseline for our real-world analysis of LGX we also choose a random direction selector for exploration. The agent takes random decisions, replicating the behavior of an \textit{`uninformed'} exploration method.

\begin{table}[t] 
    \centering
    \begin{tabular}{l c c } 
        & \multicolumn{2}{c}{RoboTHOR} \\ 
        \cmidrule(lr){2-3}
        Model & SR (\%) $\uparrow$ & SPL (\%) $\uparrow$ \\ [0.5ex] 
        \hline\hline
        CoW (FBE + CLIP) \cite{gadreCoWsPastureBaselines2022} & 15.2 & 9.7\\
        OWL-CoW (FBE + OWL-VIT) \cite{gadreCoWsPastureBaselines2022} & 27.5 & 17.2\\
        GLIP on Wheels (FBE + GLIP) & 33.2 & 20.3\\
        LGX (BLIP $\rightarrow$ LLM + GLIP) &  28.46 & 13.5\\
        \textbf{LGX (YOLO $\rightarrow$ LLM + GLIP)} & \textbf{35.0} & \textbf{21.9}\\
        \hline
    \end{tabular}
    \vspace{5.0px}
    \caption{RoboTHOR Results: Observe the improvement in Success Rate (SR) and SPL on both our approaches over the current SOTAs, CoW and OWL. }
    \label{tab:results}
    \vspace{-0.5cm}
\end{table}
 \begin{figure}[t]
     \centering
     \includegraphics[width=\linewidth]{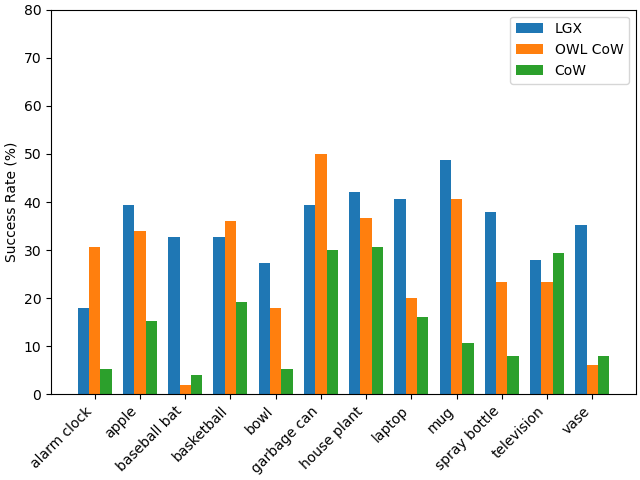}
     \caption{The class breakdown of LGX versus the OWL CoW and original CoW on RoboTHOR. LGX provides a strong improvement in localizing the baseball bat, bowl, laptop, spray bottle, and vase classes. Similar performance is noted on larger classes like television and garbage can.}
     \label{fig:thor-obj-sr}
     \vspace{-0.5cm}
 \end{figure}

\subsection{Comparison with Baselines in Simulation}
We compare the performance of our method with other models set up for the L-ZSON task in Table \ref{tab:results}. Our method significantly outperforms the OWL CoW and the original CoW with an improvement in the success rate (SR) and SPL on both. LGX also showcases an improvement in the SR over GoW. This can be attributed to 
our improved LLM-based exploration scheme on top of using GLIP for target grounding.

In Figure \ref{fig:thor-obj-sr} we compare directly with CoW and OWL across the different target objects in RoboTHOR. Our method outperformed than both baselines across smaller objects like 'bowl' and 'vase.' Our performance was similar to OWL for larger objects like 'television.' 
These results showcase the performance deficit of CoW that is likely due to the inability of CLIP to localize the target object in the image effectively.

\begin{table}[t]
    \centering
    \begin{tabular}{l c c} 
         & \multicolumn{2}{c}{RoboTHOR} \\ 
        \cmidrule(lr){2-3}
        Model & Success-Rate (\%) $\uparrow$ & PSR (\%) $\uparrow$\\ [0.5ex] 
        \hline\hline
        BLIP-Prompt & 29.2 & \textbf{100}\\
        \hline
        I-Prompt & 33.3 & 87.7\\
        Robot-Prompt & \textbf{33.8} & 71.1 \\
        Third-Person & 31.3 & 99.4\\
        \hline
        $\{O\}$first & \textbf{33.8} & 95.3\\
        Get-Closest-Object & 32.1 & 95.7\\
        "ONE word" first & 28.1 & 52.3\\
        \hline
    \end{tabular}
    \vspace{5.0px}
    \caption{Comparison of seven different prompts across three axis of change on RoboTHOR. The object-based prompts (middle and bottom) outperform than natural language-based prompts. 
    }
    \label{tab:prompt}
    \vspace{-0.5cm}
\end{table}

\subsection{Influence of Prompt Tuning Strategies}
As seen in Table \ref{tab:prompt}, the natural language-based prompts from BLIP perform worse relative to the object-based prompts despite a perfect PSR. We believe this is due to the limited action space when under the BLIP-based prompting scheme. The object-based prompts gave the LLM many different pathing options while the BLIP-based prompts were by definition associated with the four cardinal directions - potentially leading the agent towards a dead end. Additionally, we noted episodes where the LLM caused the agent to hop in a loop, continuously picking opposite directions.

No significant difference in task SR was captured over our second axis of prompt tuning denoting the perspective of the LLM relative to the robot. This is despite a wide array of PSRs for the different perspectives. The robot-perspective prompt exhibited the highest SR, but also the lowest PSR of the perspectives explored. Notably, when using the robot-perspective prompt, the LLM responded with 'no' or 'nothing' more frequently over the empty responses more commonly seen in the other prompts.

Across our changes to the structure of the prompt, there was no significant difference in SR for the object-set-first prompt or the get-closest-object prompt. However, the ``ONE word" first prompt, denoting the placement of the ``reply with ONE word" phrase before the rest of the prompt exhibited significantly worse SR and PSR. We believe this is due to the LLM no longer heeding this instruction when placed \textit{before} the remainder of the prompt. The high PSR of the get-closest-prompt indicates that picking the likely closest object may be a simpler problem for the LLM to approach. Similarly, the high PSR of the object-set-first prompt indicates that the LLM could better reference the object-set when it was placed at the beginning of the prompt.

We believe that the insignificant performance differences in task SR, despite large changes in PSR, is another indicator of RoboTHOR providing a skewed basis for this type of context dependent, intelligent exploration of the scene. 

\begin{figure}[t]
     \centering
     \includegraphics[width=\linewidth]{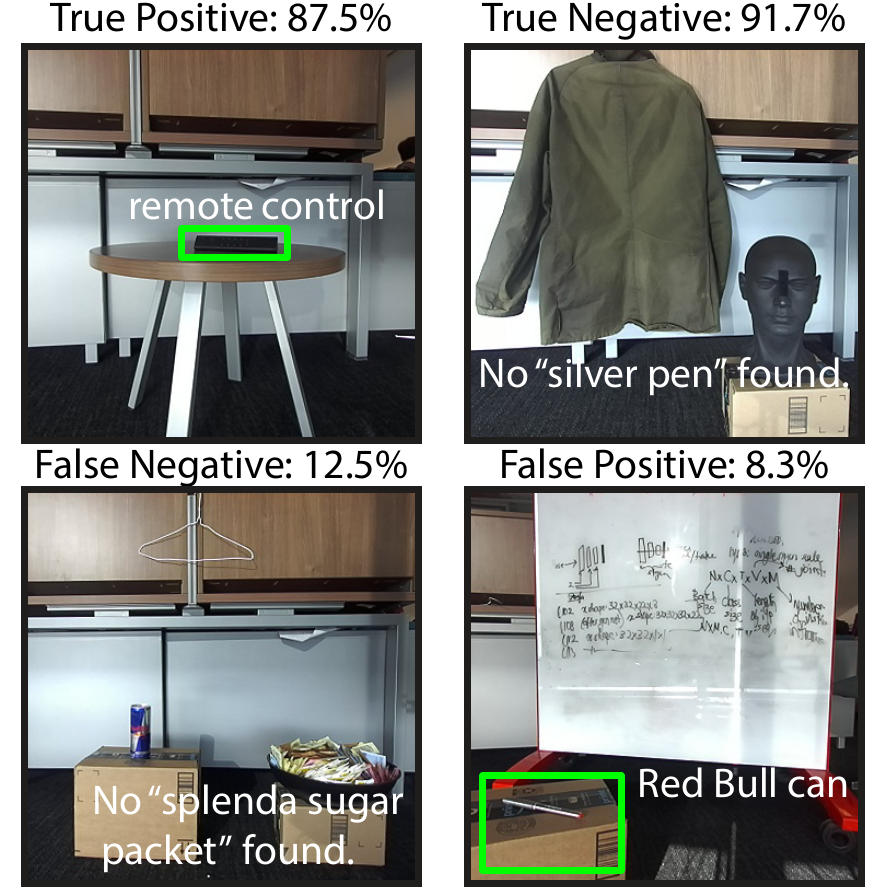}
     \caption{An sampling of GLIP success and failure cases in our real-world experimentation. When the goal object was present in the scene, GLIP accurately detected it 87.5\% of the time. Conversely, when the goal object was not present in the scene, GLIP falsely detected it 8.3\% of the time. 
     }
     \label{fig:real-world-glip}
     \vspace{-0.2cm}
 \end{figure}

\begin{table}[t] 
    \centering
    \begin{tabular}{l c} 
        \textbf{Approach (Navigation Decision $+$ Grounding)} & \textbf{Success-Rate (\%)}$\uparrow$ \\ [0.5ex] 
        \hline\hline
        Random + GLIP & 6.9\\
        GLIP-on-Wheels (GOW) & 27.8 \\
        \textbf{LGX (YOLO $\rightarrow$ LLM + GLIP)} & \textbf{54.2}\\
        \hline
    \end{tabular}
    \vspace{5.0px}
    \caption{In our real-world experimentation, our model significantly outperformed both random and GLIP-on-Wheels baselines. 
    While all three methods utilize GLIP for target object detection, neither of the baselines integrates the scene context into the exploration phases of the task.
    }
    \label{tab:real}
    \vspace{-0.4cm}
\end{table}

 \begin{figure}[t]
     \centering
     \includegraphics[width=\linewidth]{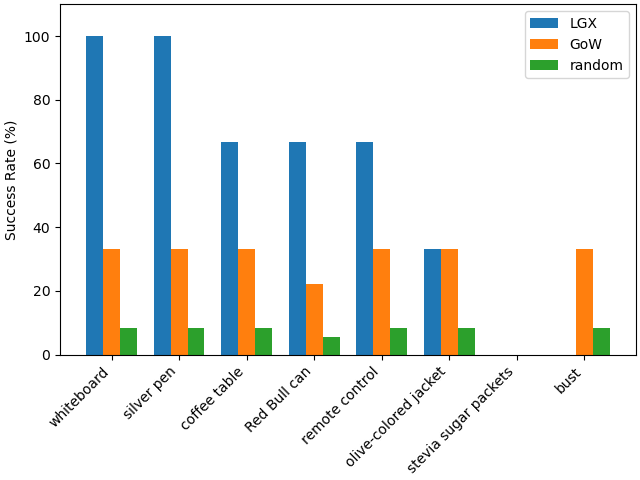}
     \caption{A comparison of individual object success rates for our method and the baselines in our real world study. Our method outperformed the baselines across the majority of targets, but notably failed to localize the 'bust' object. None of the methods could localize the 'stevia sugar packets' as GLIP failed to detect them.}
     \label{fig:real-world-obj-sr}
     \vspace{-0.5cm}
 \end{figure}

\subsection{Comparison with Baselines in the Real World}
In our real-world experiments, we consider a two-phase approach as described in the setup earlier.
As Table \ref{tab:real} indicates, LGX significantly outperforms chosen baselines, improving upon the SR of GoW by 26.4\% and the SR of Random with GLIP by 47.3\%. 
This resulted in GoW navigating to the correct room 33\% of the time while Random with GLIP explored the objects in the starting scene with the same frequency as exploring the hallway. All of the success rates were also effected by the failure cases of GLIP, specifically false negatives when attempting to detect the `stevia sugar packets' and false positives for the `Red Bull can' and the `stevia sugar packets' (see Figure \ref{fig:real-world-glip}). 

The LLM behavior in our method during our real world experimentation is characterized by three potential cases as shown in Figure \ref{fig:real-world-gpt}. The success case occurs 54.2\% of the time and is a result of the robot agent successfully navigating from the starting room into the hallway, then into the room that contains the target, before detecting the target with GLIP. 

In a Phase 1 failure case, the agent does not enter the hallway as the LLM believes one of the objects in the starting scene likely will lead to the target. One example of this we noted was when the target object is `\textit{Red Bull can}', the LLM would output `\textit{desk}' when the starting scene was the office. Although the target can is actually in the kitchen, it is plausible that it would lie on a 'desk,' explaining this output from the LLM.
In the Phase 2 failure case, the agent enters a room that does not include the target object. This occurred 20.8\% of the time with our method. This case is associated with the LLM poorly relating the target object other objects in the target room. One example of this case is the 'olive-colored jacket' which the LLM typically believed would be found near the 'desk.' 
A breakdown of the system performance for each target object is found in Figure \ref{fig:real-world-obj-sr}. Our method failed to localize the 'bust' believing it to be associated with the 'desk.' However, the relative success of the baselines indicates that GLIP succeeded in detecting the 'bust' once inside the correct room.

More details and ablations are presented in the discussion section.
\section{Limitations, Conclusions, and Future Work}
In this work we present a novel algorithm for language-based zero-shot object goal navigation. Our method leverages the capabilities of Large Language Models (LLMs) for making navigational decisions and open-vocabulary grounding models for detecting objects described using natural language. We showcase state of the art results on the RoboTHOR baseline, study the structure and phrasing of the LLM prompts that power our exploration, and validate our approach with real-world experiments. 

Our method still includes a number of failure cases, especially when the LLM incorrectly localizes the target object. Future work should explore varying the context fed to the LLM by filtering the list of objects detected or providing a history of visited objects. Similarly, exploration of which objects produced an outsized effect would be useful. Future work should also look into improving the SR and SPL metrics such that they may be more informative for zero-shot navigation tasks.

\section{Discussion}

We present some additional experiments and analyses in this section, while also answering some related questions about the real-world applicability of LGX. 

\begin{figure}[h!]
    \centering
    \includegraphics[width=\linewidth]{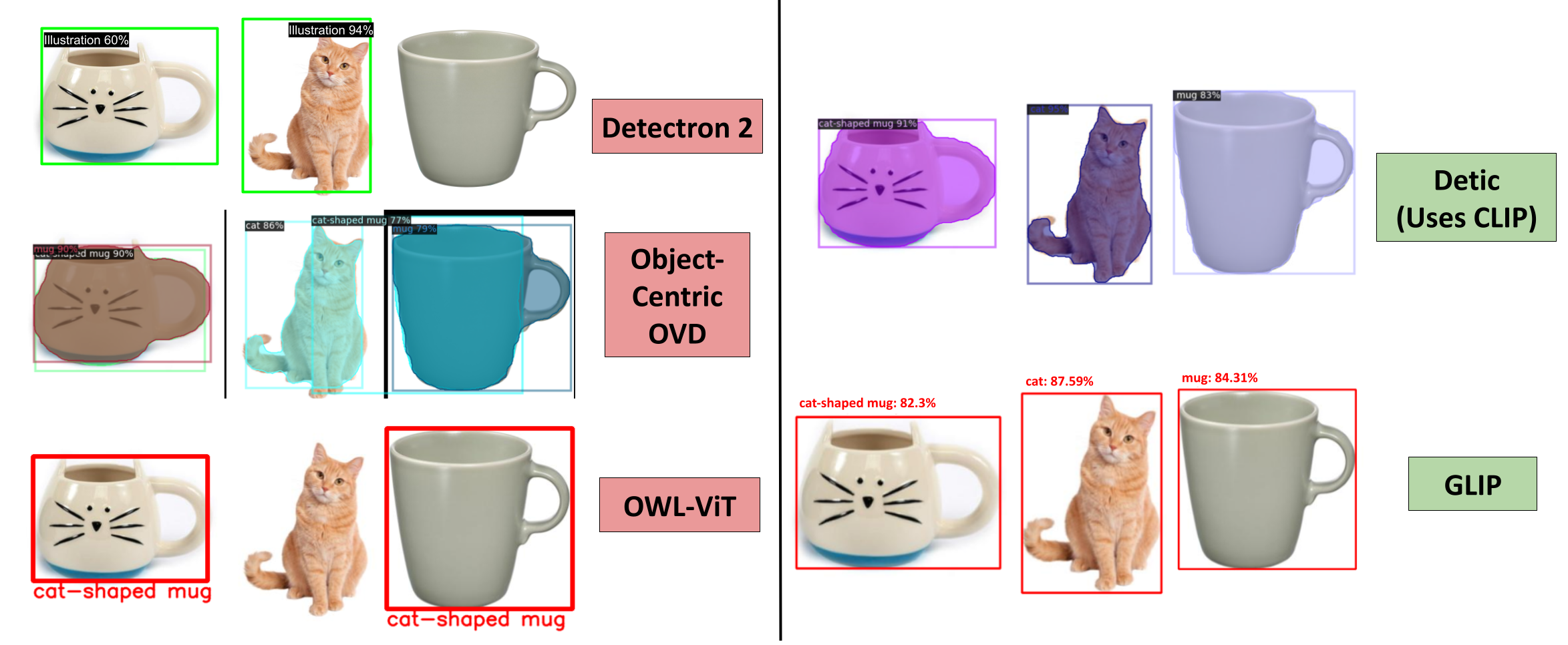}
    \caption{\textbf{Ablations on Ambiguity}: We ablate on various open-vocabulary models to assess the ambiguity of natural language when describing the object. Detectron2 \cite{detectron2}, OCOVD \cite{ocovd} and OWL-ViT \cite{owlvit} (used in OWL-CoW) show false positives and not detecting the objects sometimes. On the other hand, Detic \cite{detic} and GLIP \cite{liGroundedLanguageImagePretraining2022} work well for such open-vocabulary grounding. We decided to use GLIP for its ease of implementation, and experiments with using Detic in our framework are part of future work.}
    \label{fig:objectdetcomp}
\end{figure}

\subsection{Real-World Challenges}
Free-form natural language descriptions and partially observable environments make ObjectNav a complex task in the real-world. In zero-shot settings performance usually comes down to two main factors; \textbf{Exploration} and \textbf{Visual Object Grounding}, both of which face challenges in a real-world implementation. These are presented as follows

\begin{itemize}
    \item \textbf{Visual Object Grounding}:
    Object grounding in simulation environments is significantly more constrained than in the real world. The simulator has a set number of available classes, and sometimes characteristics of these objects are not available to ground their natural language descriptions. Users in the real world are expected to use free-form language and the agent must adapt accordingly. We attempt to overcome this challenge by using an open-vocabulary grounding model, GLIP. GLIP grounds targets from the language used in the prompt, as opposed to finding all visual correspondences, allowing for the detection of objects described in free-form natural language. We explore GLIP's performance in the second part of our real-world experimentation described in Figures \textbf{5} and \textbf{6}.  

    \item \textbf{Exploration}: 
    While exploring real-world environments, the agent is bound to a dynamically changing partially observable environment, where the entire map and associated target objects may constantly keep changing. 
    An \textit{uninformed} exploration approach such as Frontier Based Exploration (FBE) can be inefficient, since the assumption of a static environment does not exist, causing the agent to re-explore regions that it has already seen. We tackle this issue using the LLM for \textit{informed} decision-making, relying on commonsense knowledge to explore the environment. Unlike FBE which relies on selecting frontiers in free space, our LLM-based exploration is mapless, making informed choices about directions that the agent must take based on the environment surrounding it. In our real-world benchmarks, we validate the decision-making capability of the LLM via a two-phase approach presented in Section \textbf{V.A} and Figure \textbf{4}.
    
\end{itemize}

\subsection{Why GLIP?} Our approach uses GLIP \cite{GLIP}, which is an open-vocabulary object detector, which rather than proposing every object from an open-set, only detects objects mentioned in the text prompt. As an open vocabulary object detector, GLIP is known to take ambiguous text descriptions and ground those descriptions in the environment. LGX's ObjectNav performance is indeed sensitive to the performance of GLIP. 
Figure 3 in the manuscript highlights the superior capability of GLIP in tackling scenarios which are ambiguous to models without an open-vocabulary. When the user refers to a ``cat-shaped mug", GLIP is successfully able to identify this unique object from a group of images with a \textit{cat} and a \textit{mug}. We present an ablation on the performance of other SoTA zero-shot open vocabulary object detectors \emph{---}
\textit{Detectron2} \cite{detectron2},
\textit{OCOVD} \cite{ocovd}, 
\textit{OWL-ViT} \cite{owlvit},
\textit{Detic} \cite{detic} and \textit{GLIP} \cite{GLIP} in Figure \ref{fig:objectdetcomp} below.
Figure \ref{fig:objectdetcomp} qualitatively shows a positive inference in that GLIP is successfully able to tackle ambiguous situations with free-form language, reinforcing our usage for open-vocabulary grounding in LGX.




{\small
\bibliographystyle{IEEEtran}
\bibliography{refs}
}

 \addtolength{\textheight}{-8cm}   

\end{document}